\documentclass[conference]{IEEEtran}
\IEEEoverridecommandlockouts
\usepackage{cite}
\usepackage{amsmath,amssymb,amsfonts}
\usepackage{algorithmic}
\usepackage{graphicx}
\usepackage{textcomp}
\usepackage{xcolor}
\usepackage{url} 
\usepackage{booktabs}   % 三线表必备
\usepackage{multirow}   % 合并行必备
\usepackage{graphicx}   % 缩放表格必备
\usepackage[table]{xcolor} % 表格颜色 (灰色背景)
\usepackage{amssymb}    % 上下箭头符号
\usepackage{amssymb}
\usepackage{booktabs}
\usepackage{cite}
\usepackage{amsmath,amssymb,amsfonts}
\usepackage{algorithmic}
\usepackage{graphicx}
\usepackage{textcomp}
\usepackage{xcolor}
\usepackage{booktabs} % 用于三线表
\usepackage{multirow} % 用于合并行
\usepackage{graphicx}
\usepackage{amsmath}
\usepackage{amssymb}
\usepackage{bm}       % 用于粗体希腊字母（可选）
\usepackage{booktabs} % 如果有表格的话aa
\def\BibTeX{{\rm B\kern-.05em{\sc i\kern-.025em b}\kern-.08em
    T\kern-.1667em\lower.7ex\hbox{E}\kern-.125emX}}
\begin{document}

\title{FedOPAL: One-Shot Federated Learning via Analytic Visual Prompt Tuning\
}

\author{Lingyu Qiu, Daniela Annunziata, Stefano Izzo, Fabio Giampaolo, Francesco Piccialli\textsuperscript{$\dagger$} \\

Department of Mathematics and Applications “R. Caccioppoli”,\\
University of Naples Federico II, Italy\\ 

\{lingyu.qiu, daniela.annunziata, stefano.izzo, fabio.giampaolo, francesco.piccialli\}@unina.it\\
}
% \and
% \and
% \and

\maketitle
 
\renewcommand{\thefootnote}{}
\footnotetext{$\dagger$ Corresponding author.}
\renewcommand{\thefootnote}{\arabic{footnote}}

\begin{abstract}
With the widespread deployment of basic models in edge intelligence, communication bandwidth has become a core bottleneck restricting the scalability of federated learning. Although one-shot federated learning alleviates this problem by minimizing communication rounds, existing iterative fine-tuning or knowledge distillation methods still face challenges such as high server-side computational costs and hyperparameter sensitivity. Analytical federated learning achieves efficient gradient-free aggregation using least-squares closed-form solutions, but in environments with non-independent and identically distributed data, its static feature assumptions fail, leading to feature manifold misalignment and severely impairing model performance. To address this contradiction, this paper proposes the FedOPAL framework. This framework adapts the visual prompts as feature rectifiers, actively correcting the feature distribution of heterogeneous data to a linearly separable space by applying local proximal constraints, thereby satisfying the theoretical assumptions of analytical federated learning. Experimental results show that FedOPAL not only significantly outperforms the original analytical methods on several benchmarks, but also achieves accuracy comparable to state-of-the-art iterative methods while maintaining zero server-side training costs, providing a new engineering paradigm for efficient collaboration of large models on the edge. 
\end{abstract}
\newcommand\blfootnote[1]{%
  \begingroup
  \renewcommand\thefootnote{}\footnotetext{#1}%
  \endgroup
}
\blfootnote{Code is available at: \url{https://github.com/Lynn0925/FLICS}}

\section{Introduction}

Federated Learning (FL) has achieved significant success as a privacy-preserving distributed machine learning paradigm, particularly in medical image analysis\cite{adnan2022federated}, financial risk control\cite{whitmore2025privacy}, and the Internet of Things (IoT)\cite{zhang2022federated}. 
However, deep learning is currently transitioning into the era of large-scale pre-trained foundation models, leading to an exponential increase in model parameters. For instance, deploying models like ViT\cite{dosovitskiyimage} and CLIP\cite{radford2021learning} significantly increases the communication overhead between edge devices and servers. Meanwhile, relying on simple networks is no longer sufficient to meet the growing performance demands of modern industry~\cite{liu2025efficient}. In scenarios with limited bandwidth or unstable connections, such as autonomous vehicles or satellite communications, the high-frequency parameter interaction required by the traditional federated averaging algorithm FedAvg has become computationally prohibitive \cite{quan2025federated}. Therefore, realizing One-Shot Federated Learning (OFL) that can complete model aggregation with only one communication round has become a key technology to break through this bottleneck\cite{AMATO2026132088}.

To achieve single-round aggregation, existing research methods, such as those based on knowledge distillation\cite{fedsd2c,zhang2022dense,kang2023one} and data synthesis approaches\cite{Yang_Su_Li_Xue_2023}, typically employ server-side retraining. In these approaches, clients upload model parameters or generators, and the server utilizes public datasets or synthetic data to integrate knowledge via iterative optimization (e.g., ensemble distillation). While this strategy successfully circumvents multiple communication rounds, it merely shifts the computational burden from the edge to the server, rather than reducing the overall system load.

Recently, statistical computation methods based on pre-trained models have been proposed \cite{guan2025capture,he2025afl}, introducing a new paradigm for OFL. They leverage the powerful representational capabilities of pre-trained models, using methods such as least squares to compute analytical solutions at the mathematical level directly. This allows federated learning to eliminate the need for additional client training, obtaining results solely through computation. This approach provides a new perspective on one-shot federated learning; if a sufficiently powerful model is available, the final task in a single aggregation becomes how to stimulate the model's capabilities.

However, this approach relies on the strong assumption: the features extracted by the frozen backbone across heterogeneous client distributions must be high-quality and linearly separable~\cite{he2025afl}. In real-world non-IID scenarios, data distributions across different clients vary significantly, often causing the feature manifold extracted by the frozen model to be misaligned. A purely mathematical solver cannot adjust the feature extraction process itself, and its analytical solution inevitably degenerates in the face of chaotic feature statistics.

To address these challenges, we argue for a mechanism to steer the frozen backbone's feature extraction process prior to applying the linear layer, all while maintaining communication efficiency. This paper proposes FedOPAL, which utilizes Visual Prompt Tuning (VPT) to inject learnable tokens into the embedding sequence of the foundation model. Unlike pixel-level modifications, these prompts act as lightweight, learnable contexts that rectify local features and adapt them to the Analytic Federated Learning (AFL). The main contributions of this paper are as follows:

\begin{itemize}
    \item The FedOPAL framework is proposed to address the misalignment problem in statistical OFL methods. We redefine visual prompt tokens as distribution modulators in the embedding space, actively steering the frozen backbone to map heterogeneous data onto linearly separable manifolds, which ensures the effectiveness of analytical solutions under complex distributions.

    \item We utilize the closed-form solution property to replace traditional iterative fine-tuning of the Vision-Language Model with efficient algebraic operations. Under this design, the client handles feature adaptation via prompt tuning, while the server only needs to perform statistical synthesis, significantly reducing system overhead.

    \item Extensive experiments on complex benchmarks demonstrate that FedOPAL significantly outperforms existing analytical baselines in heterogeneous environments. Furthermore, our method achieves accuracy comparable to state-of-the-art iterative methods, proving that analytics-based approaches are sufficient for the practical deployment needs of large-scale edge intelligence applications.
\end{itemize}

\section{Related Work}
\subsection{One-Shot Federated Learning}
One-Shot federated learning aims to achieve model convergence within a single communication round. Early approaches primarily relied on data synthesis or knowledge distillation, where the server trains a generator to recover client data distributions~\cite{zhang2022dense, zhu2021data,song2023federated,dai2024enhancing}. Methods such as joint data-model optimization~\cite{dai2024enhancing} and dataset distillation~\cite{song2023federated} have been employed to improve distillation performance. 
Additionally, recent works have incorporated diffusion models~\cite{mendieta2024navigating}, dataset synthesis~\cite{chen2024one}, and XOR-based mixup augmentation~\cite{shin2020xor} to address privacy concerns.
However, these methods often incur high computational costs due to the iterative training of generative models.
In contrast, statistical-based methods offer a computationally efficient alternative by leveraging sufficient statistics. Instead of transmitting gradients, these methods aggregate local statistical information to derive a global closed-form solution. AFL~\cite{he2025afl} achieved gradient-free aggregation by modeling the classifier training as a distributed least squares problem. FedCGS~\cite{guan2025capture} explored combining such analytical solvers with prototype learning to enhance feature robustness. While highly efficient, these methods are strictly dependent on the alignment of feature statistics and the inherent representational power of the pre-trained model.

\subsection{Prompt Learning}
Prompt Learning initially emerged in Natural Language Processing (NLP) and was subsequently introduced to Vision-Language Models (VLMs)~\cite{zhou2022learning}. In the context of federated learning, FedOTP~\cite{li2024global} and PromptFL~\cite{guo2023promptfl} have attempted to learn prompt parameters through a federated approach~\cite{guo2023pfedprompt}. 
However, existing work largely focuses on fine-tuning text prompts. In CLIP-like architectures, the text encoder generates the weights for the linear classifier; thus, optimizing text prompts is essentially equivalent to optimizing classifier weights. Given that Analytic Federated Learning already provides a direct, closed-form solution for the optimal linear classifier, iteratively optimizing text prompts becomes inefficient and logically redundant. Conversely, Visual Prompting modifies the input embedding space (or pixels), providing a stronger capability to reshape the underlying feature manifold. To the best of our knowledge, visual prompting has not yet been applied to Analytic Federated Learning frameworks.
\section{Methodology}
\label{sec:method}
We propose the FedOPAL framework, which aims to efficiently adapt frozen foundation models to heterogeneous federated data through single-round communication. FedOPAL decouples feature alignment from classifier learning: it employs visual prompt tuning to locally rectify the feature space and leverages analytic learning (AL)\cite{he2025afl,tang2026deepafl} to achieve optimal alignment of the global classifier.

\begin{figure*}[t]
    \centering
    \includegraphics[width=1.0\linewidth]{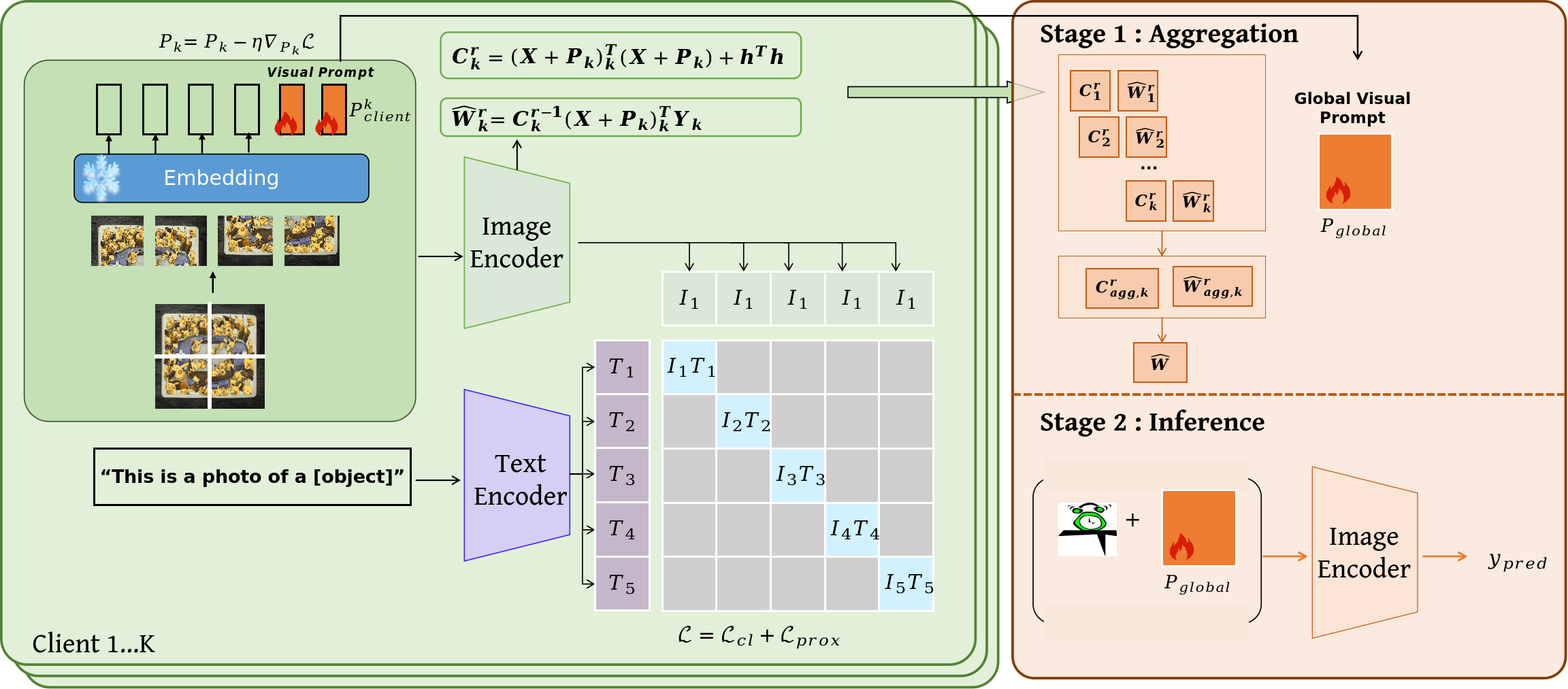}
    \caption{\textbf{Overview of FedOPAL.} (Left) In the local training phase, clients freeze the foundation model backbone and optimize a set of continuous \textit{visual prompt tokens} $\mathbf{P}_k$ inserted into the input embedding sequence. Using the rectified features extracted from the prompt-augmented sequences, clients compute closed-form sufficient statistics ($\mathbf{R}_k, \mathbf{C}_k$). (Right) The server aggregates these statistics to analytically derive the optimal global classifier $\mathbf{W}^*$ and averages the prompt tokens, achieving efficient one-shot federated learning.}
    \label{fig:framework}
\end{figure*}

\subsection{Preliminaries}

\subsubsection{Analytic Federated Learning}
We adopt the Analytic Learning (AL) framework~\cite{he2025afl} to solve the classifier learning problem in a communication-efficient manner. Consider a classification task with $C$ classes across $K$ clients. Let $\mathcal{D}_k = \{(\mathbf{x}_{i}, \mathbf{y}_{i})\}_{i=1}^{N_k}$ denote the local dataset of client $k$, where $\mathbf{y}_i \in \mathbb{R}^C$ is the one-hot label vector. The goal is to learn a linear classifier $\mathbf{W} \in \mathbb{R}^{d \times C}$ on top of a frozen feature extractor $f(\cdot)$ by minimizing the $L_2$-regularized Mean Squared Error (MSE):

\begin{equation}
    \mathbf{W}^* = \arg\min_{\mathbf{W}} \sum_{k=1}^{K} \sum_{i=1}^{N_k} \|\mathbf{W}^\top f(\mathbf{x}_{i}) - \mathbf{y}_{i}\|^2_2 + \lambda \|\mathbf{W}\|_F^2,
    \label{eq:afl_obj}
\end{equation}

where $\lambda$ is a regularization coefficient. This optimization problem has a closed-form solution derived from aggregated sufficient statistics:

\begin{equation}
    \mathbf{W}^* = \left( \sum_{k=1}^K \mathbf{R}_k + \lambda \mathbf{I} \right)^{-1} \left( \sum_{k=1}^K \mathbf{C}_k \right),
    \label{eq:closed_form}
\end{equation}

where $\mathbf{R}_k \in \mathbb{R}^{d \times d}$ and $\mathbf{C}_k \in \mathbb{R}^{d \times C}$ are the local autocorrelation and cross-correlation matrices, respectively:
\begin{equation}
    \mathbf{R}_k = \sum_{(\mathbf{x}, \mathbf{y}) \in \mathcal{D}_k} \mathbf{h} \mathbf{h}^\top, \quad \mathbf{C}_k = \sum_{(\mathbf{x}, \mathbf{y}) \in \mathcal{D}_k} \mathbf{h} \mathbf{y}^\top,
\end{equation}
where $\mathbf{h} = f(\mathbf{x})$ denotes the extracted feature vector.

\subsubsection{Visual Prompt Tuning (VPT)}
Instead of modifying input pixels or fine-tuning the entire backbone, we adopt Visual Prompt Tuning (VPT)~\cite{jia2022visual}, specifically the VPT-Shallow variant. This approach adapts frozen Vision Transformers (ViTs) by introducing a small set of $M=10$ learnable continuous tokens $\mathbf{P} \in \mathbb{R}^{M \times D}$ into the embedding space, where $D$ is the embedding dimension.

Given an input image $\mathbf{x}$, the ViT backbone first divides it into fixed-size patches and projects them into patch embeddings $\mathbf{E} \in \mathbb{R}^{L \times D}$. Let $\mathbf{x}_{cls} \in \mathbb{R}^{1 \times D}$ denote the learnable class token.
In our implementation, to preserve the spatial semantics of the pre-trained positional embeddings, we first apply position embeddings to the class token and patch embeddings. The learnable prompt tokens $\mathbf{P}$ are then inserted into the sequence:

\begin{equation}
    \mathbf{Z}_{raw} = [\mathbf{x}_{cls}, \mathbf{E}] + \mathbf{E}_{pos},
\end{equation}
\begin{equation}
    \mathbf{Z}_0 = \left[ \mathbf{Z}_{raw}^{(0)}, \mathbf{P}, \mathbf{Z}_{raw}^{(1:L)} \right],
    \label{eq:vpt_seq}
\end{equation}

where $\mathbf{E}_{pos}$ represents the position embeddings. $\mathbf{Z}_{raw}^{(0)}$ is the class token with positional information, and $\mathbf{Z}_{raw}^{(1:L)}$ are the patch embeddings with positional information. The prompt tokens $\mathbf{P}$ are inserted between the class token and the patches. The frozen Transformer blocks process this extended sequence, and the final classification feature $\mathbf{h}$ is obtained from the class token state at the last layer, denoted as $\mathbf{h} = f_\theta(\mathbf{Z}_0)_{[CLS]}$.

\subsection{The FedOPAL Framework}

FedOPAL integrates VPT with Analytic Learning to tackle the challenge of Non-IID data. The workflow consists of two main stages: Local Prompt Learning and Global Analytic Aggregation.

\subsubsection{Local Stage: Feature Rectification via VPT}
Each client $k$ initializes a local prompt matrix $\mathbf{P}_k$ and a temporary local linear head $\mathbf{W}_k$. The massive backbone model $f_\theta$ remains frozen.

\paragraph{Step 1: Local Prompt Optimization}
The client optimizes the prompt tokens $\mathbf{P}_k$ using gradient descent to adapt the frozen backbone to the local data distribution. The objective is to minimize the Cross-Entropy loss:
\begin{equation}
\begin{split}
    \min_{\mathbf{P}_k, \mathbf{W}_k} \frac{1}{N_k} \sum_{(\mathbf{x}, y) \in \mathcal{D}_k} & \mathcal{L}_{CE} \left( \text{Softmax}(\mathbf{W}_k^\top f_\theta(\mathbf{Z}_k(\mathbf{x}))_{[CLS]}), y \right) \\
    & + \frac{\mu}{2} \|\mathbf{P}_k - \mathbf{P}_{init}\|^2
\end{split}
\end{equation}
where $\mathbf{Z}_k(\mathbf{x})$ is the prompt-augmented sequence defined in Eq.~\eqref{eq:vpt_seq} using the local prompt $\mathbf{P}_k$, and $\mu$ is the proximal coefficient that controls the trade-off between local adaptation and global stability.

\paragraph{Step 2: Local Statistic Extraction}
Once the prompt $\mathbf{P}_k$ converges, we freeze it. The client then performs a forward pass using the prompt-augmented sequence to compute the sufficient statistics required for Analytic Learning:
\begin{equation}
    \mathbf{R}_k = \sum_{(\mathbf{x}, \mathbf{y}) \in \mathcal{D}_k} \mathbf{h}_i \mathbf{h}_i^\top, \quad \mathbf{C}_k = \sum_{(\mathbf{x}, \mathbf{y}) \in \mathcal{D}_k} \mathbf{h}_i \mathbf{y}_{i}^\top,
\end{equation}
where $\mathbf{h}_i = f_\theta(\mathbf{Z}_k(\mathbf{x}_i))_{[CLS]}$ is the rectified feature vector extracted from the frozen backbone guided by the optimized prompt.

\subsubsection{Server Stage: Dual Aggregation}
Clients upload their optimized prompt tokens $\mathbf{P}_k$ and statistic matrices $\mathbf{R}_k, \mathbf{C}_k$ to the server. The server performs a dual aggregation:

\paragraph{1. Prompt Aggregation}
The server averages the local prompt tokens to obtain a global prompt $\mathbf{P}_{global}$. This element-wise averaging works because the pre-trained backbone shares manifold similarities for a specific task. When all clients initialize from a shared embedding space, local prompts $\mathbf{P}_k$ act as orientation vectors, shifting heterogeneous local distributions toward task-aligned subspaces. Since the prompts operate in a shared semantic embedding space, element-wise averaging captures the consensus of visual contexts:
\begin{equation}
    \mathbf{P}_{global} = \frac{1}{K} \sum_{k=1}^{K} \mathbf{P}_k.
\end{equation}

\paragraph{2. Analytic Classifier Resolution}
Using the uploaded statistics, the server analytically computes the optimal global classifier $\mathbf{W}_{global}^*$ without iterative training. This step solves the regularized least squares problem over the aggregated feature space:
\begin{equation}
    \mathbf{W}_{global}^* = \left( \sum_{k=1}^K \mathbf{R}_k + \lambda \mathbf{I} \right)^{-1} \left( \sum_{k=1}^K \mathbf{C}_k \right).
\end{equation}

\subsection{Inference}

During inference, the client uses the downloaded global prompt $\mathbf{P}_{global}$ and classifier $\mathbf{W}_{global}^*$. A test image $\mathbf{x}_{test}$ is processed by constructing the global sequence $\mathbf{Z}_{global}$ following Eq.~\eqref{eq:vpt_seq} with $\mathbf{P}_{global}$. The final prediction is computed as:

\begin{equation}
    \hat{y} = \arg\max \left( (\mathbf{W}_{global}^*)^\top \cdot f_\theta(\mathbf{Z}_{global})_{[CLS]} \right).
\end{equation}
This ensures that the test data undergoes the same feature rectification logic learned during the distributed training phase.

\section{Experiments}

\subsection{Experimental Settings}

\subsubsection{Datasets}
In this paper, we used multiple datasets for evaluation to ensure the completeness: CIFAR-10 and CIFAR-100 (containing coarse-grained and fine-grained natural objects), SVHN (Street View House Numbers) and DTD (Describable Textures Dataset). These datasets cover a wide range of visual tasks, allowing us to test the method's adaptability to heterogeneous feature distributions and varying task difficulties.

The non-IID settings used a Dirichlet distribution ($\alpha \in \{0.01,0.1, 0.5\}$). The baseline models included FedAvg (ResNet/CLIP), the original AFL, FedCGS, and FedPFT. The backbone network of our method used CLIP (ViT-B/16).

\subsubsection{Data partition}
The widely used Dirichlet partitioning ($\alpha \in \{0.01,0.1, 0.5\}$) strategy\cite{neal2000markov} is adapted to evaluate the robustness of our method. The training set is distributed to $C$ clients according to $Dir(\alpha)$, where the parameter $\alpha$ controls the strength of data heterogeneity. This allows us to evaluate performance under various settings ranging from highly heterogeneous (small $\alpha$) to perfectly homogeneous (large $\alpha$) distributions. In this paper, unless otherwise specified, we set $C$ to 10 and $\alpha$ to 0.1.
\subsubsection{Comparison Methods}
To evaluate the effectiveness of FedOPAL, we compare our method with a range of state-of-the-art One-Shot Federated Learning (OFL) algorithms. These algorithms fall into two main categories based on their underlying methods: Traditional generation and distillation-based, and foundation model-based.
To validate the effectiveness of our proposed framework, we conducted extensive comparisons against two categories of state-of-the-art (SOTA) federated learning methods: (1) KD/Generator-based methods: Co-Boosting\cite{dai2024enhancing}, DENSE\cite{zhang2022dense} which rely on data synthesis or auxiliary data, and (2) Statistical based methods: AFL\cite{he2025afl}, FedPFT\cite{beitollahi2024parametric}, FedCGS\cite{guan2025capture}) which utilize foundation models.
\subsubsection{Model Settings and Hyperparameters.}
To leverage the power of foundational vision-language knowledge, we adopt the CLIP model with the ViT-B/16 architecture as our backbone network. The model is initialized with the official pre-trained weights. 
In contrast, following the standard procedure in their previous work, we use the backbone network specified in their paper to evaluate baseline methods (e.g., ResNet-18 or ResNet-50).

Regarding the federated training settings, we simulate the Non-IID data distribution using the Dirichlet distribution $Dir(\alpha)$, varying $\alpha$ in $\{0.01, 0.1, 1.0, 10\}$. Unless otherwise specified, we set the number of local training epochs $E=5$ for all clients to ensure sufficient local updates. The proximal coefficient $\mu$ is set to 0.1, and the analytic regularization $\lambda$ is set to $10^{-6}$ by default to ensure numerical stability during matrix inversion. The graph regularization parameter $r_g$ is set to 0. All experiments are conducted using the PyTorch framework on one A6000 NVIDIA GPU.

\begin{table}[t]
  \centering
  \setlength{\tabcolsep}{2pt}
  \caption{Comparison with KD/Generator-based methods. Results are grouped by data heterogeneity (Non-IID vs. IID).}
  \label{tab:comparison_kd_gen_ours}
    \begin{tabular}{lcccccc}
    \toprule
    \multirow{2}{*}{\textbf{Method}} & \multicolumn{3}{c}{\textbf{Non-IID} ($\alpha=0.1$)} & \multicolumn{3}{c}{\textbf{IID} ($\alpha=1.0$)} \\
    \cmidrule(lr){2-4} \cmidrule(lr){5-7} % 在两组标题下分别划线，(lr)让线稍微短一点，以区分两组
          & \textbf{CIFAR10} & \textbf{CIFAR100} & \textbf{SVHN} & \textbf{CIFAR10} & \textbf{CIFAR100} & \textbf{SVHN} \\
    \midrule
    Co-Boosting & 20.78     & 13.98     & \textbf{56.15}     & 55.28 & 22.56 & \textbf{85.67} \\
    DENSE & 40.56     & 15.84     & 42.57     & 59.98 & 30.49  & 82.03 \\
    AFL   & 82.50 & 58.56 & 53.97 & 82.50 & 58.56 & 53.45 \\
    FedPFT & 78.38 & 59.75 & 34.02 & 79.23 & 61.37 & 34.99 \\
    FedCGS & 86.11 & 64.58 & 57.45 & 86.11 & 64.58 & 57.45 \\
    \textbf{Ours} & \textbf{93.72} & \textbf{75.51} & 47.05 & \textbf{93.90} & \textbf{75.86} & 47.03\\
    \bottomrule
    \end{tabular}%
\end{table}
\begin{table}[htbp]
  \centering
  \caption{Comparison of pre-training based methods across four datasets with varying heterogeneity ($\alpha$).}
  \label{tab:comparison_pretrain}
  
  % 设置紧凑列间距
  \setlength{\tabcolsep}{4pt}

    \begin{tabular}{llcccc}
    \toprule
    \multirow{2}{*}{\textbf{Dataset}}&\multirow{2}{*}{\textbf{Method}} &  \multicolumn{4}{c}{\textbf{Data Heterogeneity}} \\
    &&$\alpha=0.01$ & $\alpha=0.1$ & $\alpha=1.0$ & $\alpha=10$ \\
    \midrule
    \multirow{4}{*}{CIFAR10} & AFL   & 82.50 & 82.50 & 82.50 & 82.50 \\
          & FedPFT & 77.47 & 78.38 & 79.23 & 78.79 \\
          & FedCGS & 86.11 & 86.11 & 86.11 & 86.11 \\
          & \textbf{Ours} & \textbf{92.19} & \textbf{93.72} & \textbf{93.90} & \textbf{92.51} \\
    \midrule
    \multirow{4}{*}{CIFAR100} & AFL   & 58.56 & 58.56 & 58.56 & 58.56 \\
          & FedPFT & 58.68 & 59.75 & 61.37 & 61.99 \\
          & FedCGS & 64.58 & 64.58 & 64.58 & 64.58 \\
          &\textbf{Ours} & \textbf{74.75} & \textbf{75.51} & \textbf{75.86} & \textbf{75.00} \\
    \midrule
    \multirow{4}{*}{SVHN} & AFL   & 53.48 & 53.97 & 53.45 & 53.27 \\
          & FedPFT & 33.20 & 34.02 & 34.99 & 35.97 \\
          & FedCGS & \textbf{57.45} & \textbf{57.45} & \textbf{57.45} & \textbf{57.45} \\
          &\textbf{Ours} & 41.48 & 47.05& 47.03 & 51.57 \\
    \midrule
    \multirow{4}{*}{DTD} & AFL   & 60.32 & 60.80 & 60.59 & 61.17 \\
          & FedPFT & 62.82 & 62.18 & 43.88 & 34.63 \\
          & FedCGS & 2.13  & 2.13  & 2.13  & 2.13 \\
          & \textbf{Ours} & \textbf{64.79} & \textbf{64.79} & \textbf{63.88} & \textbf{64.57} \\
    \bottomrule
    \end{tabular}%
\end{table}

\begin{table*}[ht]
  \centering
  \caption{\textbf{Hyperparameter Sensitivity Analysis across Heterogeneity Levels.} We evaluate the impact of the proximal coefficient $\mu$, regularization $\lambda$, and local epochs $E$ on three datasets under severe ($\alpha=0.1$) and moderate ($\alpha=1.0$) non-IID settings. The default configuration is $\mu=0.1, E=5, \lambda=0$. "$-$" indicates the default setting.}
  \label{tab:ablation_alpha_comparison}
    \begin{tabular}{l l c c c c c c}
    \toprule
    \multirow{2}{*}{\textbf{Parameter}} & \multirow{2}{*}{\textbf{Value}} & \multicolumn{2}{c}{\textbf{CIFAR-10}} & \multicolumn{2}{c}{\textbf{CIFAR-100}} & \multicolumn{2}{c}{\textbf{SVHN}} \\
    \cmidrule(lr){3-4} \cmidrule(lr){5-6} \cmidrule(lr){7-8}
     & & $\alpha=0.1$ & $\alpha=1.0$ & $\alpha=0.1$ & $\alpha=1.0$ & $\alpha=0.1$ & $\alpha=1.0$ \\
    \midrule
    \rowcolor{gray!15} \textbf{Default} & \textbf{--} & 93.72 & 93.90 & 75.51 & \textbf{75.86} & 47.05 & 47.03 \\
    \midrule
    \multirow{2}{*}{Proximal Coeff. ($\mu$)} 
      & $\mu = 0.01$ & 92.19 & 93.61 & 74.75 & 75.14 & 47.28 & 50.36 \\
      & $\mu = 0.5$ & 93.46 & 93.39 & \textbf{75.89} & 72.90 & 47.15 & 46.68 \\
    \midrule
    
    \multirow{2}{*}{Regularization ($\lambda$)} 
      & $\lambda = 10^{-6}$ & 93.74 & 93.88 & 75.85 & 75.85 & 46.94 & 46.83 \\
      & $\lambda = 1.0$ & 93.93 & \textbf{93.98} & 75.41 & 75.41 & 47.05 & 47.03 \\
    \midrule
    
    \multirow{3}{*}{Local Epochs ($E$)} 
      & $E = 1$ & 93.75 & 93.95 & 75.06 &74.87 & 46.33 & 51.72 \\
      & $E = 10$ & \textbf{93.98 }& 93.91 & 75.75 & 75.09 & 46.42 & \textbf{54.91} \\
      & $E = 20$ & 93.88 & 93.82 & 75.69 & 72.47 & \textbf{47.29 }& 54.72 \\
      
    \bottomrule
    \end{tabular}
    \label{tab:ablation_all}
\end{table*}
\subsection{Comparative Results}
In this section, to explore the efficacy of our method, we analyze the results on three aspects: (1) Evaluation on Data Heterogeneity; (2) Adaptation to the number of clients, and (3) Effects of the proposed components.

To thoroughly assess our method's capability in realistic federated scenarios, we conduct extensive experiments across varying degrees of data heterogeneity (Dirichlet $\alpha \in \{0.01, 0.05, 0.1\}$), where lower $\alpha$ values represent more extreme non-IID distributions.

To validate our framework, we conducted extensive comparisons against KD/Generator-based methods (Co-Boosting, DENSE) and Pre-training based methods (AFL, FedPFT, FedCGS). As shown in Table~\ref{tab:comparison_kd_gen_ours}, generator-based methods exhibit significant fragility in highly heterogeneous settings, with CoBoosting and DENSE failing to converge under the Non-IID setting ($\alpha=0.1$). In contrast, our method leverages the robust prior from CLIP to demonstrate exceptional stability, achieving \textbf{93.90\%} accuracy on CIFAR-10 in the IID setting, surpassing DENSE by over 50\%. This confirms that leveraging a powerful multi-modal prior is far more effective than synthesizing pseudo-samples for one-shot federated learning.

Table~\ref{tab:comparison_pretrain} further details our performance against pre-training-based SOTA methods across four datasets. Our approach achieves new SOTA performance on general object datasets; notably on CIFAR100 ($\alpha=0.01$), we outperform the strongest competitor FedCGS by 10.17\%. A striking advantage is observed on the texture-centric \textbf{dtd} dataset, where FedCGS suffers a catastrophic collapse (2.13\% accuracy) likely due to the failure of its prototypes on non-object features, whereas our method maintains a high accuracy of $\sim$64\%. While FedCGS performs slightly better on SVHN due to the domain gap between CLIP's natural-image-dominated pre-training and specific digit recognition tasks, our framework proves to be a more versatile solution overall. Furthermore, our method exhibits remarkable stability across varying heterogeneity levels, with negligible performance fluctuation (e.g., 0.32\% variance on CIFAR10), effectively disentangling model performance from client data distributions.

Regarding the results on the SVHN dataset, we observed that FedOPAL performs slightly worse than FedCGS. This is primarily due to the domain difference between CLIP's pre-training and the specific digit domain. Because FedOPAL uses a frozen backbone network, its shallow cues have limited generalization to out-of-distribution digit features. However, FedOPAL maintains a significant advantage on all other natural image and texture datasets (CIFAR, DTD), demonstrating superior generalization ability.
\subsection{Ablation Study and Parameter Sensitivity}

In this section, we conduct extensive ablation analysis on the CIFAR-100 dataset under a non-independent and identically distributed (non-IID) setting ($\alpha=0.1$) to investigate the impact of key hyperparameters and design choices in FedOPAL. The results are summarized in Table \ref{tab:ablation_all}.

\subsubsection{Impact of Proximal Coefficient $\mu$}

The proximal coefficient $\mu$ in FedOPAL controls the tradeoff between local optimization and global stability. It acts as a regularization term, dominating the bias between local visual cues and global anchors. To investigate its impact, we vary the value of $\mu$ within the range $\{0.01, 0.1, 1.0\}$.
As shown in Table \ref{tab:ablation_all}, the performance exhibits an inverted U-shaped trend. When $\mu \to 0$, the constraint disappears, leading to overfitting of local cues to sparse local data and causing feature misalignment during aggregation. Conversely, excessively large $\mu$ (e.g., $\mu=1.0$) overly restrict the adaptability of the prompts, preventing them from effectively correcting the feature distribution. The best performance is achieved near $\mu \in [0.1, 1.0]$, indicating that moderate constraints can effectively balance local feature correction and global semantic consistency.

\subsubsection{Impact of No-IID Distribution ($\alpha$)} To evaluate FedOPAL's robustness to statistical heterogeneity, we conducted experiments with varying degrees of data distribution skew. The concentration parameter $\alpha$ of the Dirichlet distribution ranges from $\{0.05, 0.1, 0.5, 1.0, \infty\}$, where smaller $\alpha$ indicates a greater degree of label imbalance among clients. Due to the violation of the static feature assumption, the performance of standard AFL drops sharply as $\alpha$ decreases, while FedOPAL maintains relatively stable accuracy even under extreme heterogeneity ($\alpha=0.01$). This demonstrates that visual cues successfully correct distorted local feature manifolds into a unified global space, making the analytical solution effective even when the client data is highly non-independent and identically distributed.

\subsubsection{Impact of AFL Regularization ($\lambda$)} The regularization parameter $\lambda$ in the analytical solution $\mathbf{W}^* = (\mathbf{R} + \lambda \mathbf{I})^{-1}\mathbf{C}$ in AFL ensures numerical stability during matrix inversion, especially when the aggregated autocorrelation matrix $\mathbf{R}$ is ill-conditioned. We examine the model's sensitivity to $\lambda$ by scanning the value of $\lambda$ from $10^{-6}$ to $10^{1}$.
Experimental results show that FedOPAL is highly robust to the choice of $\lambda$. On all three datasets (CIFAR-10, CIFAR-100, and SVHN), the impact of changing the $\lambda$ parameter on performance is generally within 0.5\%. This insensitivity suggests that the feature matrices extracted by the frozen backbone are inherently well-conditioned and discriminative. The feature space optimized by VPT is statistically stable enough that the global autocorrelation matrix $\sum \mathbf{R}_k$ does not suffer from rank deficiency, even without explicit regularization. Consequently, FedOPAL simplifies the hyperparameter tuning process, as the default setting ($\lambda \approx 0$) is sufficient for achieving optimal performance in most scenarios. 
\section{Discussion and Conclusion}This paper proposes a framework called FedOPAL, which optimizes the efficiency of analytical federated learning (AFL) on non-independent and identically distributed (Non-IID) data by introducing visual cues as feature correctors. FedOPAL combines local fine-tuning with global analytical aggregation, achieving state-of-the-art (SOTA) classification performance while maintaining strict single-pass communication and zero server-side training cost. This work not only reveals the deep mathematical connection between visual cues and analytical learning but also provides a robust and efficient solution for deploying base models in edge computing environments.

\section*{Acknowledgment}
This work is partially supported by the project: TUAI: Towards an Understanding of Artificial Intelligence via a transparent, open and explainable perspective, Marie Curie Doctoral Network, ID: 101168344.
This work is partially supported by the project: FLINT - Prog. n.: F/380134/01-03/X77 - MIMIT

\bibliographystyle{ieeetr}
\bibliography{flics}

\end{document}